\documentclass[conf]{new-aiaa}
\usepackage[utf8]{inputenc}
\usepackage[margin=1in]{geometry}
\usepackage{tikz}
\usepackage{makecell}
\usepackage{sectsty}
\usepackage{titlesec}
\usepackage[parfill]{parskip}
\usepackage[font={small}]{caption}
\usepackage{todonotes}
\usepackage{graphicx}
\usepackage{amsmath} 
\usepackage{subcaption}
\usepackage{ifpdf}
\graphicspath{{./images/}}

\makeatletter
\renewcommand*\env@matrix[1][*\c@MaxMatrixCols c]{%
  \hskip -\arraycolsep
  \let\@ifnextchar\new@ifnextchar
  \array{#1}}
\makeatother

\sectionfont{\centering\bfseries\fontsize{11}{0}\selectfont}
\subsectionfont{\centering\bfseries\fontsize{11}{0}\selectfont}
\titlespacing{\section}{0pt}{5mm}{0mm}
\titlespacing{\subsection}{0pt}{5mm}{0mm}

\title{\textbf{ 
Assistive Relative Pose Estimation for On-orbit Assembly using Convolutional Neural Networks 
}}

\author{Shubham Sonawani\footnote{ PhD student at ASU, sdsonawa@asu.edu}}
\affil{Interactive Robotics Laboratory, Arizona State University, Tempe, AZ, 85281, USA}

\author{Ryan Alimo\footnote{Lead Machine Learning Scientist in Deep Learning Technologies group at Jet Propulsion Laboratory (JPL), sralimo@jpl.nasa.gov}, 
Renaud Detry\footnote{Research Scientist in Mobility and Robotics Division at Jet Propulsion Laboratory (JPL), renaud.j.detry@jpl.nasa.gov},
Daniel Jeong\footnote{Intern at Jet Propulsion Laboratory (JPL), daniel.p.jeong@jpl.nasa.gov}, 
Andrew Hess\footnote{Intern at Jet Propulsion Laboratory (JPL), andrew.hess@jpl.nasa.gov}
}
\affil{Jet Propulsion Laboratory, California Institute of Technology, Pasadena, CA, 91109, USA}

\author{Heni Ben Amor\footnote{ Professor of School of Computing, Informatics, and Decision Systems Engineering, hbenamor@asu.edu}}
\affil{ Interactive Robotics Laboratory, Arizona State University, Tempe, AZ, 85281, USA }
\date{}

\newcommand{\paren}[1]{\left(#1\right)}
\renewcommand{\vec}[1]{\mathbf{#1}}

\newcommand{\drawquadx}[7] {
\draw[fill=#7] 	(#1, #2, #3) -- (#1, #2, #3 + #6) -- (#1, #2 + #5, #3 + #6)	-- (#1, #2 + #5, #3) -- cycle;
}

\newcommand{\drawquady}[7] {
\draw[fill=#7]	(#1, #2, #3) -- (#1 + #4, #2, #3) -- (#1 + #4, #2, #3 + #6) -- (#1, #2, #3 + #6) -- cycle;
}

\newcommand{\drawquadz}[7] {
\draw[fill=#7]	(#1, #2, #3) -- (#1 + #4, #2, #3) -- (#1 + #4, #2 + #5, #3) -- (#1, #2 + #5, #3) -- cycle;
}

\newcommand{\block}[7] { % x, y, z, dx, dy, dz, color
\drawquadx{#1 + #4}{#2}{#3}{#4}{#5}{#6}{#7}
\drawquady{#1}{#2 + #5}{#3}{#4}{#5}{#6}{#7}
\drawquadz{#1}{#2}{#3 + #6}{#4}{#5}{#6}{#7}
}

\newcommand{\conv}[3] {
\drawquadx{#1 + \dx}{#2}{#3}{\dx}{\dy}{\dz}{\convcolor}
\drawquady{#1}{#2 + \dy}{#3}{\dx}{\dy}{\dz}{\convcolor}
\drawquadz{#1}{#2}{#3 + \dz}{\dx}{\dy}{\dz}{\convcolor}
}

\newcommand{\pool}[3] {
\drawquadx{#1 + \dx}{#2}{#3}{\dx}{\dy}{\dz}{\poolcolor}
\drawquady{#1}{#2 + \dy}{#3}{\dx}{\dy}{\dz}{\poolcolor}
\drawquadz{#1}{#2}{#3 + \dz}{\dx}{\dy}{\dz}{\poolcolor}
}

\newcommand{\dense}[3] {
\drawquadx{#1 + \dx}{#2}{#3}{\dx}{\dy}{\dz}{\densecolor}
\drawquady{#1}{#2 + \dy}{#3}{\dx}{\dy}{\dz}{\densecolor}
\drawquadz{#1}{#2}{#3 + \dz}{\dx}{\dy}{\dz}{\densecolor}
}

\newcommand{\merge}[3] {
\block{#1}{#2}{#3}{\dx}{\dy}{\dz}{\mergecolor}
}
\newcommand{\quatinput}[3] {
\drawquadx{#1 + \dx}{#2}{#3}{\dx}{\dy}{\dz}{\quatcolor}
\drawquady{#1}{#2 + \dy}{#3}{\dx}{\dy}{\dz}{\quatcolor}
\drawquadz{#1}{#2}{#3 + \dz}{\dx}{\dy}{\dz}{\quatcolor}
}

\begin{document}
\maketitle
\begin{abstract}
  Accurate real-time pose estimation of spacecraft or object in space is a key capability necessary for on-orbit spacecraft servicing and assembly tasks. Pose estimation of objects in space is more challenging than for objects on Earth due to space images containing widely varying illumination conditions, high contrast, and poor resolution in addition to power and mass constraints. In this paper, a convolutional neural network is leveraged to uniquely determine the translation and rotation of an object of interest relative to the camera. The main idea of using CNN model is to assist object tracker used in on space assembly tasks where only feature based method is always not sufficient. The simulation framework designed for assembly task is used to generate dataset for training the modified CNN models and, then results of different models are compared with measure of how accurately models are predicting the pose. Unlike many current approaches for spacecraft or object in space pose estimation, the model does not rely on hand-crafted object-specific features which makes this model more robust and easier to apply to other types of spacecraft. It is shown that the model performs comparable to the current feature-selection methods and can therefore be used in conjunction with them to provide more reliable estimates.  
\end{abstract}
% \fontsize{8}{8}\selectfont
% \textbf{\textit{Abstract}--Accurate real-time pose estimation of spacecraft or object in space is a key capability necessary for on orbit spacecraft servicing and assembly tasks. Pose estimation of objects in space is more challenging than for objects on Earth due to space images containing widely varying illumination conditions, high contrast, and poor resolution in addition to power and mass constraints. In this paper, a convolutional neural network is leveraged to uniquely determine the translation and rotation of an object of interest relative to the camera. The main idea of using CNN model is to assist object tracker used in on space assembly tasks where only feature based method is always not sufficient. The simulation framework designed for assembly task is used to generate dataset for training the modified models and, then results are compared with different models with notion of how accurately models are predicting the pose. Unlike many current approaches for spacecraft or object in space pose estimation, the model does not rely on hand-crafted object-specific features which makes this model more robust and easier to apply to other types of spacecraft. It is shown that the model performs comparable to the current feature-selection methods and can therefore be used in conjunction with them to provide more reliable estimates.}

\fontsize{10}{12}\selectfont

\section{Introduction}
A key ability necessary for on-orbit spacecraft servicing is accurate estimation of the position and orientation of the spacecraft in real time, as this allows the servicing spacecraft to automatically fine-tune its trajectory and timing. Similarly, such technology is useful for close-formation flying \cite{hadaegh-2001}, precision formation flying (PFF) \cite{hadaegh-2003, hadaegh-2004}, active debris removal, and distributed space systems  with planetary science applications \cite{damico-2017, matsuka-2019}. In such missions, particularly in on-orbit assembly application, the vision-based sensors are beneficial to  estimate the pose of neighboring objects, or spacecraft \cite{sharma-2018,capuano-2019}, while the target object is known, but uncooperative. 
Using monocular vision-based camera in space for navigation purposes has gained interest during the recent years \cite{sharma-2018a,pedro-2019,capuano-2019a,capuano-2020,damico-2020} since these sensors are low-power, low-cost, small, and accessible particularly in small satellites and cubesats. In these settings, we want to determine the relative pose (position and attitude) of a target object with respect to the chaser (i.e., camera). In this paper, we focus on relative pose estimation using vision-based camera for a known, but an uncooperative object. 

The relative pose estimation is a well-studied field in many terrestrial pose-estimation tasks particularly for pose estimation of indoor objects.  A major challenge in the terrestrial pose-estimation tasks is clutter and object occlusions \cite{pose cnn}. However, relative pose estimation of objects in space is a different problem from pose determination of objects on Earth. 
This issue is not as pronounced in space, where visible foreground typically contains a single object, possibly accompanied by a few background distractors such as planets or stars. In-space visual conditions are however more challenging than conditions met on Earth: because of the lack of atmosphere, light diffusion is entirely absent in space. The lack of diffusion creates much stronger shadows: object surfaces are either exposed to the full power of incident sunlight, or receive almost no light at all, resulting in extreme image contrast. Additionally, space hardware is impeded by technological constraints such as radiation resilience, power consumption and mass limits, which impact image resolution, sensor noise, and computational resources. Considering prior limitations, it is highly important to use passive sensing to do pose estimation in space and, also the active sensors such as light detection and ranging sensor (LIDAR) and RADAR have large masses and are power hungry which makes active sensing not suitable for space applications with power constraints.

On-orbit assembly task comprised of autonomous state estimation and manipulation of external objects in quasi-static environment. Due to previously discussed computational and power constraints, pose estimation using passive sensors such as the monocular vision sensor is important technology for mission critical tasks. In addition to this, this technology will be important feat for future missions such as Phoenix program by DARPA\cite{phoenix darpa} and The Restore-L servicing mission by NASA\cite{restore L}. Recent efforts shows demonstration of assembly task using monocular vision based tracking and highly occluded object \cite{surp,tracker} as proof of concept.
However, tracker came short in initialization phase to instantly localize the object in image frame and lock on to it for further tracking. Inaccurate initialization is highly susceptible due to an object occlusion in features based tracking techniques. Instead of completely relying on features based approach, a backup CNN based pose estimation can be used as corrective mechanism to feature based tracking during initialization phase. In this paper, we propose two transfer learning based models for pose estimation from monocular images. The models are not specific to a single type of spacecraft or to a object of interest and can thus be applied to other models of object by training on a sufficiently large dataset of the new object models. %Our main contribution is modifying a convolutional neural network (CNN) designed for object recognition by replacing the max pooling layers with parallel  layers to improve the model's accuracy on satellite pose estimation.

Our main contribution is modifying and improving accuracy of convolutional neural network, initially designed for object recognition, to do a relative pose estimation of known object. It is trained on synthetic dataset generated using simulation testbed shown in fig.\ref{fig:realmodel}.
%  \todo[inline]{Ryan1: pose estimation challenge and why we need CNN in a better way. 
%  }
It has been shown recently that Convolutional Neural Networks (CNN) that were trained solely on synthetic data exhibit improved performance on actual spaceborne images compared to existing onboard feature-based algorithms \cite{tracker}. Specifically, CNN are more robust to adverse illumination and dynamic Earth background in spaceborne images. However, this has only been shown empirically to date and there is no explanation on why an estimator may fail and in some situations may fail. 
The structure of this paper is as follows. Following this introduction, Section \ref{sec:relatedworks} reviews some of the recent works in relative pose estimation for on-orbit assembly tasks, and Section \ref{sec:problemstatement} describes the problem that we are trying to solve. Section \ref{sec:dataset} describes the synthetic data generated for this study. Section \ref{sec:method} explains the design and implementation of loss functions and CNN model. Later, Section \ref{sec:results} shows results  obtained from dataset compared to other models. In the end, Section \ref{sec:conclusions}, draws some conclusions and discusses future works.
\begin{figure}[h!]
    \centering
    \includegraphics[scale = 0.5]{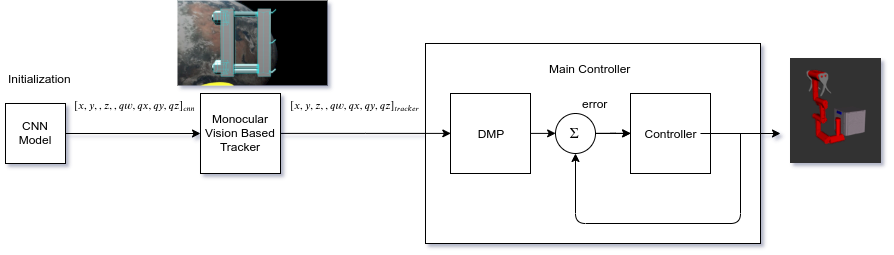}
    \caption{Assembly framework}
    \label{fig:assembly_framework}
\end{figure}

% Ryan change and update the descriptions 
\subsection{ Relative Pose Estimation for On-orbit Assembly } \label{sec:relatedworks}
% \todo[inline]{ Shubham and Renuad: We need a literature review for On orbit assembly.}

\begin{figure}[h!]
    \centering
     \begin{subfigure}{0.25\textwidth}
    \centerline{
    \includegraphics[scale=0.35]{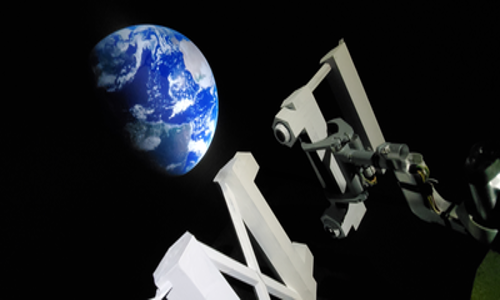}}
      \caption{Realworld testbed experiment}
      \label{fig:real}
    \end{subfigure}
    \hfil
    \begin{subfigure}{0.25\textwidth}
    \centerline{
    \includegraphics[scale=0.35]{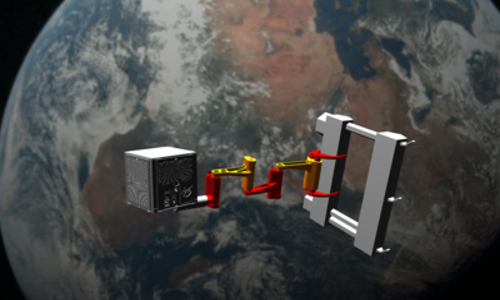}}
    \caption{Simulated testbed experiment}
     \label{fig:sim}
    \end{subfigure}
    \caption{Demonstration of the robotic arm performing on-orbit assembly in simulated and realworld testbed with space like lighting condition using vision based feedback controller. For more information about this experiment see \cite{surp}.  }
    \label{fig:realmodel}
\end{figure}

On-orbit assembly of spacecraft or structures in space has been proposed several times in past considering factors such as bigger size of assembled structure, deployment risk etc. In addition to this, space robotics gained a lot of attention in last decade because of improvement in low compute and space grade hardware. In may 2015, as per \cite{roadmaps} NASA updated many technology area such as Robotics and Autonomous Systems, Human Exploration Destination systems etc. with goal of having In orbit assembly technology \cite{Zimpfer}\cite{Doggett_r}. Previously, assembly of space system using robotic technology was demonstrated in NASA Langley Research Center \cite{langley}. However, demonstration involved complex vision sensing for successful completion of task, however make it nearly impossible to implement on low compute power hardware. Nonetheless, In recent work, arm augmented cubesat \cite{remora} was designed by JPL to showcase hardware capability to perform autonomous In orbit assembly. Cubesat with arm uses on board passive sensing to obtain relative pose of object of interest, in this case that object is a truss. Later this obtained relative pose can be used to do visual servoing based control of robotic arm for grasping and manipulation of object. By leveraging this design, authors have designed simulation testbed shown in fig. \ref{fig:sim}. More information about this testbed is given in \ref{sec:dataset} and assembly framework to perform on-orbit assembly using relative pose estimation can be visualized in fig. \ref{fig:assembly_framework}%more information about it can be found \cite{simulation}

% Can you check what needs to fixed once?
%  I mostly went over paper. let me know

\subsection{Problem Statement}\label{sec:problemstatement}
This paper proposes a solution to solve tracking problem using CNN based pose initialization. In this problem, More emphasis is given on designing a CNN model for a relative pose estimation. Dataset used for training the CNN model is generated from simulation of assembly testbed. Considering promising application of such CNN model based pose estimation in future mission critical tasks, empirical study of trained model is necessary. Thus, authors have analyzed the trained CNN network for good and worst predictions in order to understand which part of the network learns better. Specifically, by looking at heatmaps of convolutional layers we can see where exactly network is focusing for predictions similar to \cite{deep_features}. %The qualitative analysis of CNN model is important to understand how network    
Furthermore, In assembly task, pose estimation and tracking of the object in real time is critical information for proper manipulation and grasping. Also, Tracking might get lost due to temporary object occlusion from arm itself or partial visibility in camera frame. In that case, reinitialization of tracking by estimating object pose instantly and latching onto the object is important task. Future work will be on integrating this model with tracker to test the performance of assembly task on realworld testbed with framework as shown in fig. \ref{fig:assembly_framework}

In our previous work, \cite{surp} we demonstrated ability to perform assembly task using monocular vision based tracking and remora arm \cite{remora} designed by JPL. This work shows successful real-time decision making using vision based feedback and dynamic motion primitives on low compute platform. Simulation test bed designed for this work is used to generate synthetic dataset.However, for this study generating a set of images that are similar to the real-world can be challenging and costly and is out of scope of this work. There are a few places in academic institutions that have such a capability such as Stanford's SLAB \cite{slab}, Caltech's CAST \cite{cast} that have such a capability. For this work, we use synthetically generated dataset that simulates testbed with on the orbit conditions shown in fig. \ref{fig:realmodel}. In future work, realworld images of object will be generated instead of synthetic using designed testbed shown in fig. \ref{fig:real} The following section describes the synthetic dataset generated for this work.
\section{Dataset}\label{sec:dataset}

\begin{figure}[htb]
    \centering % <-- added
\begin{subfigure}{0.25\textwidth}
  \includegraphics[width=\linewidth]{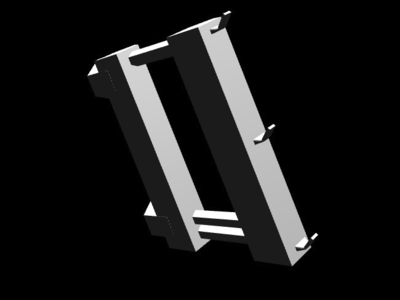}
%   \caption{Image 1}
  \label{fig:org_img1}
\end{subfigure}\hfil % <-- added
\begin{subfigure}{0.25\textwidth}
  \includegraphics[width=\linewidth]{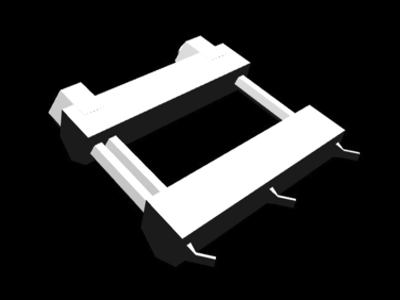}
%   \caption{Image 2}
  \label{fig:org_img2}
\end{subfigure}\hfil % <-- added
\caption{Synthetic images obtained from simulation test with 3D pose of object wrt camera frame}
\medskip

\begin{subfigure}{0.25\textwidth}
  \includegraphics[width=\linewidth]{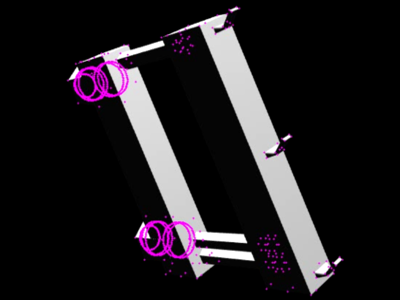}
%   \caption{Validation of image 1}
  \label{fig:reproj_1}
\end{subfigure}\hfil % <-- added
\begin{subfigure}{0.25\textwidth}
  \includegraphics[width=\linewidth]{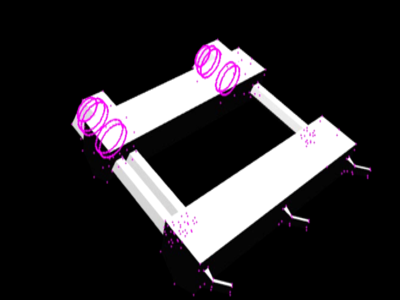}
%   \caption{Validation of image 2}
  \label{fig:reproj_2}
\end{subfigure}\hfil % <-- added
\caption{Verification of obtained 3D pose of object in image by reprojecting vertices of object on to the camera frame}
% 	\caption{ Illustration of synthetic data used in this study. }
	\label{fig:challenges}
\end{figure}

Training a convolutional neural network models usually require large number of training data of hundred thousand images e.g., \cite{coco,openimage}.  However, generating real-world images from space can be challenging for the application of on-orbit assembly. As a result, in this paper, we validate our framework using synthetic dataset generated using simulation testbed discussed previously. Fig. \ref{fig:challenges} shows a few synthetically generated images and corresponding validated images.
Our dataset is generated leveraging a physic-based simulation platform, named Gazebo \cite{gazebo}, within Robot Operating System (ROS)\cite{ros}. This simulation testbed is capable of simulating on-orbit physics condition such as quasi-static behaviour of object, illumination changes, rigid body dynamics in space. In addition to this, it can simulate variety of sensors such as monocular camera, stereo camera, LIDAR sensor and actuators which makes it all in one framework to get dataset and perform experiments. 

 In this dataset, a truss shaped object is used with asymmetric shape to avoid different labelling to images consisting same object with 180 degrees rotation making two images to look exact copy of each other. Thus, asymmetric shape of object helps to avoid having any skewed data samples in generated dataset. Using simulated/synthetic camera, translation and orientation of object of interest, in this case truss, was obtained using careful manipulation in simulated in-space lighting conditions. Furthermore, considering in space assembly task using cubesat, size of truss is approximately 0.38$\times$0.2$\times$0.05 (meters) and rotated around camera frame within reachable distance for arm. Images were captured initially in resolution of 720$\times$1280 pixels where each image was labelled with pose of truss with respective to camera frame. Ones dataset is generated, it is verified using reprojecting the labelled poses of truss on to the image frame. Since simulation allows access to synthetic camera's intrinsic parameters, labelled pose is transformed with respect to camera frame. Later vertices of object obtained from 3D cad model are transformed to new pose, calculation details are shown belowfor more information. By obtaining reprojected images, validity of dataset can be confirmed as shown in \ref{fig:challenges}. 

\begin{center}
$\begin{bmatrix}
\mathbf{X_{img}} \\
\mathbf{Y_{img}} \\
\mathbf{Z_{img}} \\ 
\end{bmatrix}
= \mathbf{K} [\mathbf{I}\|\mathbf{0}] \times
\begin{bmatrix}
\mathbf{R} & \mathbf{t}\\
\mathbf{0} & \mathbf{1}
\end{bmatrix}
\times 
\begin{bmatrix}
\mathbf{X_{world}}\\
\mathbf{Y_{world}}\\
\mathbf{Z_{world}}
\end{bmatrix}$
\end{center}

\begin{center}
    \begin{tabular}{@{}l@{\ }l}
    Where, & \\
    & $\mathbf{K}$ is intrinsic matrix of camera\\
    & $\mathbf{X_{img}},\mathbf{Y_{img}},\mathbf{Z_{img}}$ are coordinates in camera frame  \\
    & $\mathbf{R}$ and $\mathbf{t}$ are rotation and translation of object\\
    & $\mathbf{X_{world}},\mathbf{Y_{world}},\mathbf{Z_{world}}$ are coordinates of object's vertices
%\text{Projection Matrix} \times \text{Transformation Matrix} \times \text{Pose of Vertices}
\end{tabular}
\end{center}

%%%
The models were tested on dataset mentioned above. The images in dataset were scaled down to a size of 224x224 pixels to be consistent with ImageNet \cite{imagenet} because learning from ImageNet was transferred to each model. Each image had a corresponding label consisting of seven numbers representing the position and orientation of the satellite with respect to the camera. The position is given by the distance in meters along the x, y, and z axes while the rotation is specified by the other four numbers, which make a unit quaternion. In order to constrain the model such that it predicts only unit quaternions, the predicted quaternions were converted to unit quaternions by dividing each component by the quaternion's magnitude both when calculating the loss during training and when determining the model's overall accuracy during testing. Quaternions were chosen over rotation matrices and Euler angles to represent the orientation of the satellite because quaternions have less redundancy than rotation matrices while still avoiding gimbal lock, which Euler angles are highly susceptible to.

\section{Relative Pose Estimator Models } \label{sec:method}
Two similar CNN models were constructed to estimate the satellite pose. Similar to \cite{sharma cnn}, transfer learning was used with weights obtained from training on the ImageNet dataset \cite{imagenet}. Conversely, the models in this paper are based on VGG-19 \cite{vgg} rather than AlexNet \cite{alexnet} due to the superior classification accuracy of VGG-19 on the ImageNet dataset. Since the task is to estimate seven continuous numbers rather than perform the 1000-way classification of ImageNet, the last layer of each of the models was replaced by a 7-node layer with no activation function.

The models use transfer learning with most of the weights frozen for three main reasons. The first is that since ImageNet and the satellite or space related dataset are somewhat similar in that they consist of color images of one main object, transfer learning can provide a good initialization to speed up training by transferring some of the knowledge of how low-level features are extracted and combined into more complicated features. The second reason is that the satellite or space related dataset is relatively small and freezing many of the weights so that they are not updated during training helps limit overfitting the training data. The final reason is because \cite{yosinski} showed that transfer learning can boost performance even after the original weights have had significant fine-tuning through training on the new dataset.

\begin{figure*}
\begin{tikzpicture}[xscale=0.071, yscale=0.060]
\def \convcolor {cyan!20}
\def \poolcolor {gray!20}
\def \densecolor {orange!50}
\def \mergecolor {green!40}
\def \dx{2} \def \dy{30} \def \dz{40}

% Block 1
\conv{0}{0}{0}
\conv{4}{0}{0}
\def \dx{2} \def \dy{25} \def \dz{35}
\pool{8}{0}{0}

% Block 2
\def \dx{4} \def \dy{25} \def \dz{35}
\conv{16}{0}{0}
\conv{22}{0}{0}
\def \dx{4} \def \dy{20} \def \dz{30}
\pool{28}{0}{0}

% Block 3
\def \dx{6} \def \dy{20} \def \dz{30}
\conv{38}{0}{0}
\conv{46}{0}{0}
\conv{54}{0}{0}
\conv{62}{0}{0}
\def \dx{6} \def \dy{15} \def \dz{25}
\pool{70}{0}{0}

% Block 4
\def \dx{8} \def \dy{15} \def \dz{25}
\conv{82}{0}{0}
\conv{92}{0}{0}
\conv{102}{0}{0}
\conv{112}{0}{0}
\def \dx{8} \def \dy{10} \def \dz{15}
\pool{122}{0}{0}

% Block 5
\def \dx{10} \def \dy{10} \def \dz{15}
\conv{136}{0}{0}
\conv{148}{0}{0}
\conv{160}{0}{0}
\conv{172}{0}{0}
\def \dx{10} \def \dy{5} \def \dz{10}
\pool{184}{0}{0}

% The end layers
\def \dx{2} \def \dy{10} \def \dz{20}
\merge{200}{0}{0}
\def \dx{2} \def \dy{2} \def \dz{30}
\dense{206}{0}{-10}
\dense{212}{0}{-10}
\def \dx{2} \def \dy{2} \def \dz{5}
\dense{218}{0}{5}

% The branch
\def \dx{4} \def \dy{15} \def \dz{25}
\pool{70}{-30}{0}
\def \dx{2} \def \dy{2} \def \dz{20}
\dense{82}{-25}{0}

% Draw the arrows for the branch.
\draw[->] (24, -12, 0) .. controls +(-90: 22cm) and +(180: 10cm) .. (58, -32, 0) {};
\draw[->] (86, -32, 0) .. controls +(0: 30cm) and +(-90: 25cm) .. (194, -12, 0) {};

\matrix at (current bounding box.north) {
	\node [circle, label=right:Convolution, fill=\convcolor] 		(conv)				{};
	\node [circle, label=right:Max Pooling, fill=\poolcolor] 		(pool)	[right of=conv, xshift=2cm]	{};
	\node [circle, label=right:Concatenate, fill=\mergecolor] 	(merge)	[right of=pool, xshift=2cm]	{};
	\node [circle, label=right:Fully Connected, fill=\densecolor](dense)	[right of=merge, xshift=2cm] {};\\
};
\end{tikzpicture}
\caption{This diagram shows the branched version of VGG-19. The branch was added to the model in order to preserve some feature-position information that was discarded by later max pooling layers.}
\label{fig:branched_model}
\end{figure*}
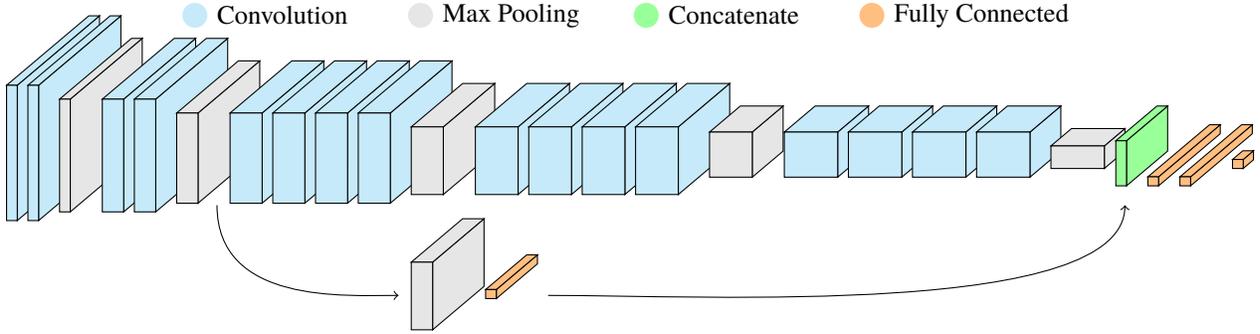

\begin{figure*}[ht]
\centering
\begin{tikzpicture}[xscale=0.050, yscale=0.060]
\def \quatcolor {purple!30}
\def \transcolor {yellow!50}
\def \convcolor {cyan!20}
\def \poolcolor {gray!20}
\def \densecolor {orange!50}
\def \mergecolor {green!40}
\def \dx{2} \def \dy{30} \def \dz{40}

\quatinput{-8}{0}{0}

% Block 1
\conv{-4}{0}{0}
\conv{0}{0}{0}
\conv{4}{0}{0}
\def \dx{2} \def \dy{25} \def \dz{35}
\pool{8}{0}{0}

% Block 2
\def \dx{4} \def \dy{25} \def \dz{35}
\conv{16}{0}{0}
\conv{22}{0}{0}
\def \dx{4} \def \dy{20} \def \dz{30}
\pool{28}{0}{0}

% Block 3
\def \dx{6} \def \dy{20} \def \dz{30}
\conv{38}{0}{0}
\conv{46}{0}{0}
\conv{54}{0}{0}
\conv{62}{0}{0}
\def \dx{6} \def \dy{15} \def \dz{25}
\pool{70}{0}{0}

% Block 4
\def \dx{8} \def \dy{15} \def \dz{25}
\conv{82}{0}{0}
\conv{92}{0}{0}
\conv{102}{0}{0}
\conv{112}{0}{0}
\def \dx{8} \def \dy{10} \def \dz{15}
\pool{122}{0}{0}

% Block 5
\def \dx{10} \def \dy{10} \def \dz{15}
\conv{136}{0}{0}
\conv{148}{0}{0}
\conv{160}{0}{0}
\conv{172}{0}{0}
\def \dx{10} \def \dy{5} \def \dz{10}
\pool{184}{0}{0}

% The end layers
\def \dx{2} \def \dy{10} \def \dz{20}
\merge{200}{0}{0}
\def \dx{2} \def \dy{2} \def \dz{30}
\dense{206}{0}{-10}
\dense{212}{0}{-10}
\def \dx{2} \def \dy{2} \def \dz{5}
\dense{218}{0}{5}

% The branch
\def \dx{4} \def \dy{15} \def \dz{25}
\pool{70}{-30}{0}
\def \dx{2} \def \dy{2} \def \dz{20}
\dense{82}{-25}{0}

% Quat
\def \dx{2} \def \dy{30} \def \dz{40}

% \transinput{-8}{-70}{0}

% Block 1
\conv{-4}{-70}{0}
\conv{0}{-70}{0}
\conv{4}{-70}{0}
\def \dx{2} \def \dy{25} \def \dz{35}
\pool{8}{-70}{0}

% Block 2
\def \dx{4} \def \dy{25} \def \dz{35}
\conv{16}{-70}{0}
\conv{22}{-70}{0}
\def \dx{4} \def \dy{20} \def \dz{30}
\pool{28}{-70}{0}

% Block 3
\def \dx{6} \def \dy{20} \def \dz{30}
\conv{38}{-70}{0}
\conv{46}{-70}{0}
\conv{54}{-70}{0}
\conv{62}{-70}{0}
\def \dx{6} \def \dy{15} \def \dz{25}
\pool{70}{-70}{0}

% Block 4
\def \dx{8} \def \dy{15} \def \dz{25}
\conv{82}{-70}{0}
\conv{92}{-70}{0}
\conv{102}{-70}{0}
\conv{112}{-70}{0}
\def \dx{8} \def \dy{10} \def \dz{15}
\pool{122}{-70}{0}

% Block 5
\def \dx{10} \def \dy{10} \def \dz{15}
\conv{136}{-70}{0}
\conv{148}{-70}{0}
\conv{160}{-70}{0}
\conv{172}{-70}{0}
\def \dx{10} \def \dy{5} \def \dz{10}
\pool{184}{-70}{0}

% The end layers
\def \dx{2} \def \dy{10} \def \dz{20}
\merge{200}{-70}{0}
\def \dx{2} \def \dy{2} \def \dz{30}
\dense{206}{-70}{-10}
\dense{212}{-70}{-10}
\def \dx{2} \def \dy{2} \def \dz{5}
\dense{218}{-70}{5}

% The branch
%\def \dx{4} \def \dy{15} \def \dz{25}
%\pool{70}{-100}{0}
%\def \dx{2} \def \dy{2} \def \dz{20}
%\dense{82}{-95}{0}

% Final output
\def \dx{2} \def \dy{2} \def \dz{10}
\dense{232}{-40}{0}

% Draw the arrows for the branch.
\draw[->] (24, -12, 0) .. controls +(-90: 22cm) and +(180: 10cm) .. (58, -32, 0) {};
\draw[->] (86, -32, 0) .. controls +(0: 30cm) and +(-90: 25cm) .. (194, -12, 0) {};

% Draw the arrows for the branch.
%\draw[->] (24, -82, 0) .. controls +(-90: 22cm) and +(180: 10cm) .. (58, -102, 0) {};
%\draw[->] (86, -102, 0) .. controls +(0: 30cm) and +(-90: 25cm) .. (194, -82, 0) {};

% Draw the arrows to final output.
\draw[->] (216, -6, 0) -- (228, -40, 0) {};
\draw[->] (218, -69, 0) -- (228, -45, 0) {};

\end{tikzpicture}
\caption{Our parallel CNN architecture for pose estimation, based on VGG-19. The colors indicate different processing stages in the network: bounded image input layer (purple), full-sized image input layer (yellow), convolution layers (blue), maxpooling layers (gray), concatenation layers (green), and fully connected layers (orange).}
\label{fig:cnn_arch_parallel}
\end{figure*}
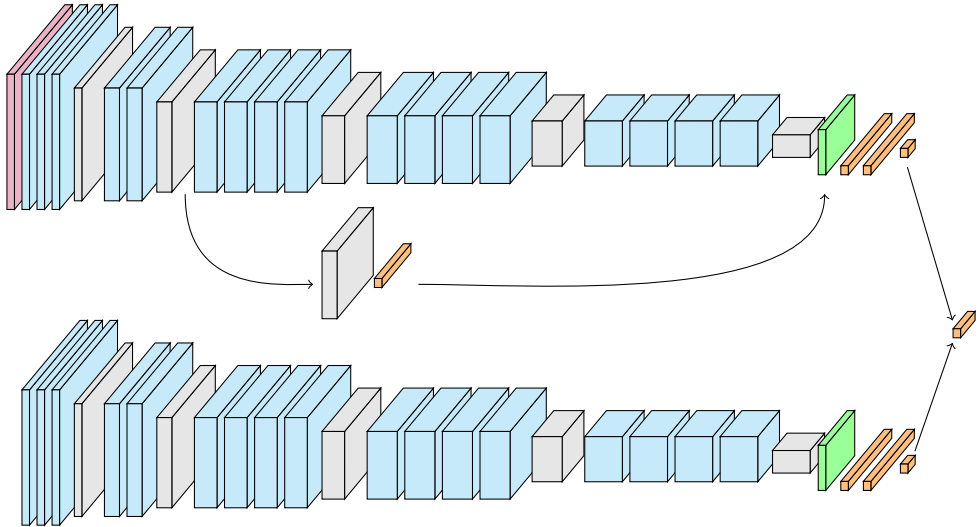

One key distinction between object classification and pose estimation is what the model should be robust against. Object classification models should predict the same labels even if the objects are slightly rotated or moved, while pose estimation models need to produce a slightly different output. Conversely, both types of models should be robust against random noise, different backgrounds, and varying illumination conditions. In many CNN architectures for object recognition, the robustness against small translations is in part due to the max pooling layers. Both models created for this work were designed to limit the feature location information lost due to the translational and rotational robustness of VGG-19. The architecture of the two models developed for this paper are shown in fig. \ref{fig:branched_model} and fig. \ref{fig:cnn_arch_parallel}.

% The branched model.

% \paragraph{Branched Model}
We used different architectures and introduced two models as Branched model, and Parallel model to determine the pose of the object.
The first model, named Branched model, used some of the feature position information to the fully connected layers at the end of the model by creating a parallel branch. In order to constrain the model to a reasonable size, the branch started after the second max pooling layer of VGG-19 and consisted of another 2x2 max pooling layer followed by a fully connected layer with 1024 nodes. The branch was merged with the rest of the model directly before the first fully connected layer at the end via concatenation. Merging the branches through concatenation was chosen over other methods such as addition because doing so did not significantly increase the number of parameters in the model and allowed for a more direct incorporation the branch's features. The architecture of this model is illustrated in Figure \ref{fig:branched_model}.

The other model introduced is called parallel model and its architecture illustrated in fig. \ref{fig:cnn_arch_parallel} shows the model with two parallel streams of the VGG network for relative distance and VGG-branched for attitude estimation of the object. This model takes in two different types of input, first one is normal image for translation prediction with 3 output nodes at the bottom of fig. \ref{fig:cnn_arch_parallel} and second one is at top of figure with input as bounded image and having branched structure from second convolution box concatanated to final max polling layer with 4 output  nodes after two fully connected layers.

\subsection{Loss Function}\label{sec:lossfunction}
% \todo{we need to use classification loss.}
The models were trained to simultaneously minimize the translational and rotational loss. The loss function used for translational error was the Euclidean distance between the predicted and actual translations. For actual and predicted translations $\vec T$ and $\vec t$, the loss is:
$$\mathcal{L}_T = \left|\left| \vec T - \vec t\right|\right|_2$$

The rotational loss is the rotational difference between the actual and predicted quaternions. By regressing to minimize the rotational difference between the quaternions rather than regressing to the actual quaternion, the models bypass the complication that the negative and positive of a unit quaternion represent equivalent rotations. For quaternions $\vec q_1$ and $\vec q_2$ with real parts $k_1$ and $k_2$ and imaginary parts $\vec v_1$ and $\vec v_2$, the real part of $\vec q_1 \vec q_2$ is given by,
$$k = k_1k_2 - \vec v_1 \cdot \vec v_2$$
The rotational difference between $\vec q_1$ and $\vec q_2$ can then be computed as,
$$\theta = 2\cos^{-1}(k)$$
Since the range of arccosine is $[0, \pi)$, then $\theta\in[0, 2\pi)$. However, since the absolute value of the angle between $\vec q_1$ and $\vec q_2$ is being minimized and for an angle $\phi$, we know $-\phi$ is equivalent to $2\pi - \phi$, each computed $\theta$ was modified to:
$$\theta = \min(\theta, 2\pi-\theta)$$

For the actual and predicted rotation quaternions $\vec Q$ and $\vec q$, the rotational loss is:
$$\mathcal L_R = 2\min\paren{\cos^{-1}\paren{\textrm{Re}\paren{\vec Q \vec q}}, 2\pi - \cos^{-1}\paren{\textrm{Re}\paren{\vec Q \vec q}}}$$

Work by \cite{geometric loss} found that training a model to regress to the position and rotation simultaneously performed better than a model that estimated them separately. Due to this, the two losses were combined into the total loss using a scaling factor $\beta$. The total loss is thus:
$$\mathcal L = \mathcal L_T + \beta\mathcal L_R$$

Adding the losses due to translational and rotational errors in this manner was done in \cite{posenet}, which found that the optimal value of $\beta$ was such that $\mathcal L_T$ and $\beta\mathcal L_R$ were roughly equal at the end of training. In order to keep $\mathcal L_T$ and $\beta\mathcal L_R$ roughly equal, $\beta$ was set to 10 for each of the tested models.

\begin{table*}
\centering
\begin{tabular}{|l|>{\centering}p{30mm}|>{\centering}p{30mm}|>{\centering}p{30mm}|}
\hline
Model & Mean Rotation Error (degrees) & Median Rotation Error (degrees) & Mean Translation Error (meters)\tabularnewline
\Xhline{1pt}
% ResNet-50 		   	& 9.26 	  & 7.41          & \tabularnewline\hline
VGG-16 			  	& 13.56 	  & 12.33 & 0.18\tabularnewline\hline
VGG-19 			 	& 12.21 	  & 11.34          & 0.15 \tabularnewline\hline
Branched VGG-19 	& 8.34 	  & 7.34          & \textbf{0.06} \tabularnewline\hline
Parallel VGG-19  	& \textbf{4.61} 	& 4.41 & \textbf{0.03} \tabularnewline\hline
\end{tabular}

\caption{This table shows the accuracy of parallel VGG-19 on the synthetic  dataset after training for 100 epochs.}
\label{results}
\end{table*}

\begin{figure}[h]
    \centering
    \includegraphics[width=0.7\textwidth]{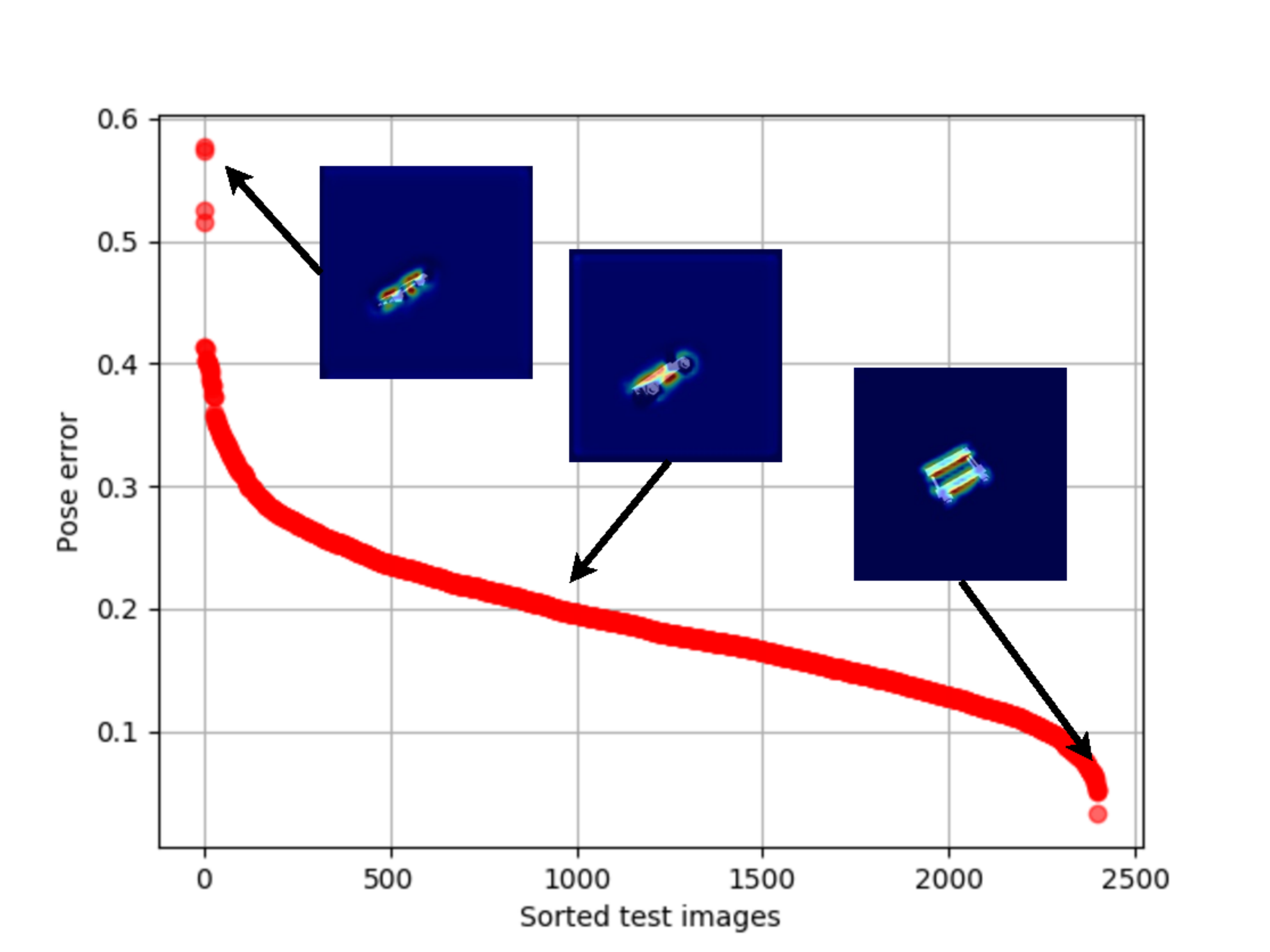}
    % need to fix caption??
    \caption{Pose error with parallel VGG-19 for sorted images with corresponding heatmaps}
    \label{fig:error}
\end{figure}

\begin{figure}[h]
    \centering
    \begin{subfigure}{0.2\textwidth}
    \centerline{
      \includegraphics[scale=0.6]{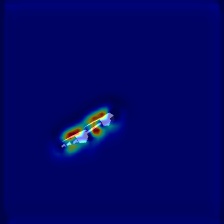}}
    \end{subfigure}
    \hfil
    \begin{subfigure}{0.2\textwidth}
      \centerline{
      \includegraphics[scale=0.6]{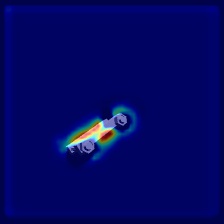}}
    \end{subfigure}
    \hfil
    \begin{subfigure}{0.2\textwidth}
    \centerline{
      \includegraphics[scale=0.6]{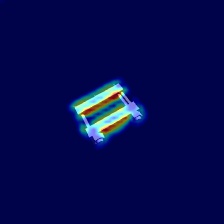}}
    \end{subfigure}
    \caption{Heatmaps of test images with highest error from left to lowest error on right obtained from Parallel VGG-19 model}
    \label{fig:heatmaps}
\end{figure}

\section{Results}\label{sec:results}
% \todo[inline]{R->Shubham: please summarize the results and observations. }
The models were tested on the pose estimation dataset described above with 80\% of the data used for training and the remaining 20\% used for testing. The training and testing data was split randomly, but each model was trained using the same data. To provide baseline metrics for the created models, several object recognition models were minimally altered and evaluated on the dataset. The baseline models were created by taking well-performing image classification models on the ImageNet dataset and replacing the last fully connected layer with a fully connected layer of seven nodes and no activation function. Additionally, transfer learning was used with weights obtained by training on the ImageNet dataset and only the weights of the fully connected layers at the end of each model were updated during training. The results are displayed in Table \ref{results}.

Results show transfer learning on original VGG16 and VGG19 model with 100 epochs gives high pose errors compared to branched VGG19 and parallel VGG19. For parallel VGG19 model performs better than other models specifically for attitude predictions. Bounded images used in attitude prediction focuses on object of interest which interestingly turned out to successful in extracting orientation information of object. On the other hand branched VGG19 and Parallel VGG19 model performed well on translation prediction of object. It shows using complete image, containing object and spacial background, is sufficient enough to obtain better translation prediction.

\begin{figure}[h!]
    \centering
    \includegraphics[width=1\textwidth]{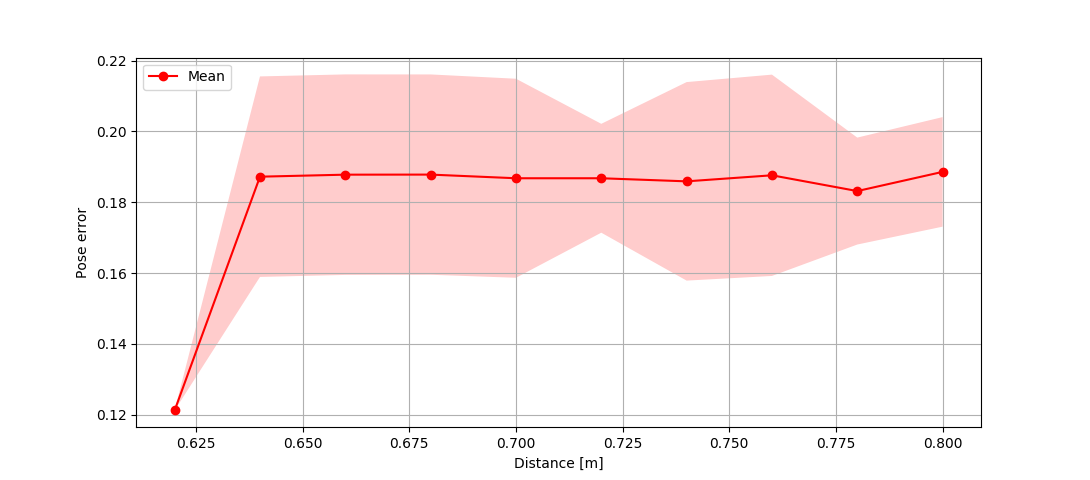}
    % need to fix caption??
    \caption{Mean and 1 standard deviation of error in prediction for parallel VGG-19 over range of distance }
    \label{fig:prediction_var}
\end{figure}

As shown in fig. \ref{fig:error}, error in prediction is higher when object is further from camera frame which can be seen in heatmaps and visible geometry of object is close to symmetric. On the other hand, error tends to be marginal or low when object is closer to camera frame and geometric features of object are dominant which seems to improve prediction confidence of network as seen in heatmaps shown in \ref{fig:heatmaps}. From graph \ref{fig:prediction_var} we can see the variation of mean pose (translational and rotational) error vs euclidian distance from camera frame. This showcases that when object is closer error remains smaller and plateaus till maximum distance. However we can see standard deviation of error is remains constant till 0.725 meters and then varies again in form of blob showing increase in error deviation as object moves further from camera frame. 

% Theses results shows 

\section{Conclusion and Future Work}\label{sec:conclusions}
The results in Table \ref{results} show that the branched model performed significantly better than the unmodified VGG-16 and VGG-19 model. The improvement over standard VGG-19 shows that the addition of the parallel branch successfully provided the end layers with features that were useful for fine-tuning the pose estimate. Moreover, the parallel  model significantly outperformed each of the other models on this dataset. This indicates that replacing the max pooling layers with parallel layers by increasing the convolution stride provides feature location information that was discarded by max pooling layers. 

The comparison between the parallel  model and the feature based \cite{sharma cnn} model reveal that a CNN model can perform comparable to feature based methods. As such, the parallel  model could be used on its own, to provide an initial guess for feature based models, or in an ensemble with feature based algorithms. One significant advantage of CNN models over feature based models is that they can be adapted for different objects by simply training the model with a sufficiently large dataset containing the new object. Conversely, feature based models require the exact structure of the object, which may not be known for certain types of objects such as asteroids.

As future work, the parallel  model should be applied to more challenging satellite pose estimation datasets and thoroughly compared against both feature based models and other CNN models. Since VGG-19 was designed for object classification rather than pose estimation, other architectures could be developed using parallel  in place of max pooling to further improve results. Additionally, the current loss function has a hyper-parameter $\beta$, which could be tuned to produce more accurate results. Furthermore, CNN model can be used in combination with feature based tracking method during initialization phase to improve results of pose estimation. This can be tested on space grade compute platform and less diffused lighting condition which can give understanding on how much model assist tracker to regain the tracking on object in simulated space conditions. Further analysis can be done on model in terms of number of deep and convolution layers to get fast inference. Authors will integrate this scheme into estimation work and apply this work to the dataset from real mission by extending the sim2real framework.

\section*{Acknowledgments}
The authors gratefully acknowledge funding from Jet Propulsion Laboratory, California Institute of Technology, under a contract with the National Aeronautics and Space Administration (NASA) in support of this work. The authors thank  Vincenzo Capuano, Kyunam Kim, Kasra Yazdani, Kingson Man, and Jonathan Chu for the fruitful discussions; also, Amir Rahmani, David Hanks, Adrian Stoica, Navid Dehghani,  Leon Alkalai,  and Fred Hadaegh for their supports.

\clearpage

\onecolumn

\end{document}